\documentclass[AMA,STIX1COL]{WileyNJD-v2}
\usepackage{moreverb}
\usepackage{ amssymb }
\usepackage{gensymb}

\newcommand\BibTeX{{\rmfamily B\kern-.05em \textsc{i\kern-.025em b}\kern-.08em
T\kern-.1667em\lower.7ex\hbox{E}\kern-.125emX}}

\articletype{Article Type}%

\received{<day> <Month>, <year>}
\revised{<day> <Month>, <year>}
\accepted{<day> <Month>, <year>}

\begin{document}

\title{Overcoming Limited Battery Data Challenges: A Coupled Neural Network Approach}

\author[1]{Aniruddh Herle}

\author[1]{Janamejaya Channegowda}

\author[2]{Dinakar Prabhu}

\authormark{Herle \textsc{et al}}

\address[1]{\orgdiv{Department of Electrical and Electronics Engineering}, \orgname{M. S. Ramaiah Institute of Technology
}, \orgaddress{\state{Karnataka}, \country{India}}}

\address[2]{\orgdiv{Department of Management Studies}, \orgname{Indian Institute of Science}, \orgaddress{\state{Karnataka}, \country{India}}}


\corres{Janamejaya Channegowda, MSRIT Post, \\ M S Ramaiah Nagar, MSR Nagar, Bengaluru, Karnataka 560054, India \email{bcjanmay.edu@gmail.com}}


\abstract[Abstract]{The Electric Vehicle (EV) Industry has seen extraordinary growth in the last few years. This is primarily due to an ever increasing awareness of the detrimental environmental effects of fossil fuel powered vehicles and availability of inexpensive Lithium-ion batteries (LIBs). In order to safely deploy these LIBs in Electric Vehicles, certain battery states need to be constantly monitored to ensure safe and healthy operation. The use of Machine Learning to estimate battery states such as State-of-Charge and State-of-Health have become an extremely active area of research. However, limited availability of open-source diverse datasets has stifled the growth of this field, and is a problem largely ignored in literature. In this work, we propose a novel method of time-series battery data augmentation using deep neural networks. We introduce and analyze the method of using two neural networks working together to alternatively produce synthetic charging and discharging battery profiles. One model produces battery charging profiles, and another produces battery discharging profiles. The proposed approach is evaluated using few public battery datasets to illustrate its effectiveness, and our results show the efficacy of this approach to solve the challenges of limited battery data. We also test this approach on dynamic Electric Vehicle drive cycles as well.}


\maketitle 
\section{Introduction}

Lithium Ion batteries (LIBs) are more relevant than ever before for meeting energy demands in the modern day. Apart from the environmental benefits of using these rechargeable batteries over conventional fossil fuels \cite{Stampatori2020}, these batteries are very energy dense and inexpensive. This makes them an ideal energy source for a variety of applications, from powering smartphones \cite{Lee2020}$^,$ \cite{Singh2020} to Electric Vehicles (EVs).\\
\indent The intersection of Machine Learning and Battery technology has become a very active area of research over the last few years. Deep Learning has seen a lot of interest for various tasks such as State-of-Health (SOH) \cite{Shen2020}$^{-}$\cite{Khumprom2019} and State-of-Charge (SOC) \cite{Hannan2020}$^{-}$\cite{How2019a} estimation. Conventional battery modeling requires extensive battery parameterization by domain experts \cite{Tran2020a}$^{-}$\cite{Marquis2019}, whereas deep learning models have the ability to automatically learn the most relevant battery features along with transient dynamics. Although these methods show a lot of promise, and seem to have the potential to become the dominant method for SOC and SOH estimation, there is a major roadblock that must first be overcome. The accuracy of any machine learning algorithm is highly correlated with the quality and quantity of training data\cite{Dubarry2020}. A model trained on a limited amount of data is unable to accurately capture all the relevant battery dynamics. In this scenario, data augmentation techniques provide a viable solution to increase the training data. Limited data challenges are largely ignored in literature, and this work aims to elucidate the problem and offer a simple and novel solution.\\
\indent Most papers that suggest new and improved machine learning based battery parameter estimation models use confidential proprietary data. This data is very rarely made open-source, essentially stifling the research in this area. There is no legitimate way to compare different approaches in two research papers if the underlying data used to train the models are different. Moreover, battery manufacturers routinely generate vast quantities of data during quality control tests, and this data is also not made available to researchers due to privacy and confidentiality reasons. There is thus a dire data scarcity of diverse battery datasets within publicly available battery data, which is needed to foster growth in this area of research.\\
\indent The use of Markov Chains for the purpose of augmenting battery dynamics has also been explored in Pyne et al. \cite{Pyne2019}. This method is akin to simulating the underlying battery electrochemistry behavior to generate adequate data for training purposes. They report that their model is able to capture the battery capacity fade over time. Most research related to data-augmentation are based around statistical and probabilistic approaches. For example, in Kamthe et al. \cite{Kamthe2021}, copula theory was used to formulate a synthetic data generation methodology. About three million voltage vs. capacity curves were generated using the approach introduced in Dubarry et al. \cite{Dubarry2020}. The authors used a mechanistic approach to capture the most vital battery dynamics for diagnostics and prognosis purposes. The problem with these approaches is that they require extensive battery parameter estimation testing, which is quite expensive and tedious.\\ 
\indent The use of Generative Adversarial Networks (GANs) \cite{Goodfellow2014} for the purpose of data augmentation has also become very popular, \cite{DosSantosTanaka2019}$^{-}$\cite{Yoon2019} especially in the field of image data augmentation \cite{Shorten2019}. GANs have been known to suffer from convergence issues \cite{Wu2018}, and typically only manage to capture limited range of the time series variations (maximum and minimum values) but fail to capture the intrinsic dynamics of the data. Moreover, training of GANs are extremely resource intensive, as it requires both the Generator and Discriminator networks to be trained together thus rendering them ineffective for resource constrained remote devices.\\ 
\indent In this paper, we introduce a novel method of data augmentation for time-series data. This method can be used in data scarce scenarios, and helps to significantly increase the amount of training data. We use the popular Oxford Battery Degradation Dataset \cite{oxford_dataset} to train and test our synthetic data generator, and then test the model architecture on the US06 and BJDST drive cycles of the Center for Advanced Life Cycle Engineering dataset \cite{Zheng2016} to evaluate the effectiveness of the proposed approach.\\

The primary contributions of the current work are listed here:
\begin{itemize}
\item This is the first work to use two simple Feedforward Neural Networks to generate battery synthetic data
\item This is the first time a coupled neural network approach has been employed to produce synthetic charge/discharge data to overcome limited data challenges
\item  This is the first work to use computationally simple neural networks which can be employed in resource constrained remote devices
\end{itemize}

\indent The paper is organized as follows: fundamentals of Deep Neural Networks are provided in Section 2, details of the methodology introduced in this paper is given in Section 3, dataset details are provided in Section 4, hyperparameter tuning is covered in Section 5. The results obtained are illustrated in Section 6, followed by conclusion in Section 7.

\section{Deep Neural Networks}

\indent A neural network is made up of several layers, and each layer has multiple neurons. These neurons are essentially the computational units of the entire model. They consist of weights and biases. An input to the neuron undergoes specific computations based on the values of these parameters, and the result is the input for the subsequent layer. The output of the last layer, which can be a regression output or a classification label, is the final output of the neural network. \\
\indent The neural network has a pre-defined error also known as the loss function. The goal of training a neural network is to minimize this loss function over the sample space. In our work, the Mean Squared Error shown in Eqn. 1 is considered as the loss function. A popular algorithm called Adam \cite{Kingma2015} is used to iteratively reduce the loss function, as it is computationally efficient. The loss is back-propagated and adjustments are made to the values of weights and biases until an optimal model is obtained. After training is completed satisfactorily, the model can be deployed to predict the target variable in a real-life setting. It essentially behaves like a function approximator.
\begin{equation}
    \mathcal{L}_{MSE} = \frac{1}{n} \Sigma^n _{i=1} (\hat{y_i} - y_i)^2
\end{equation}
\indent In Eqn. 1, $n$ is the number of samples in a minibatch,$y_i$ is the input vector and $\hat{y_i}$ is the output vector at the $i$-th instance.\\
\indent Traditional Neural Network approaches use one or two computational layers. These are usually referred to as Artificial Neural Networks (ANNs). Extending this idea, we can add more layers to the mix, resulting in a deeper model architecture called Deep Neural Networks (DNN). The capability of these architectures to learn non-linear relations among the inputs and outputs is greatly enhanced. Simply put, a multilayer stack of neural layers constitutes a deep learning model. In principle, a Deep Neural Network with enough layers can approximate any function. In this article, we use the traditional Feedforward Neural Network. In this network, information flows from the input layer to the output layer (forward direction).

\section{Methodology}

\indent In this article, we confine the discussion to Feedforward Neural Networks. The dataset used has three battery parameters: Voltage, Temperature and Charge Capacity (which is converted to State-of-Charge). We formulate the Artificial Sequence Generation task as follows:
\begin{enumerate}
\item Two separate neural networks are trained for each battery parameter
\item One neural network is trained to generate a Discharging profile, and another neural network is trained to produce a Charging profile
\end{enumerate}

For each battery parameter, it is also possible to couple the two neural networks together sequentially to produce a large amount of synthetic battery parameter data.\\

\indent For the sake of brevity, we will use the following nomenclature throughout this work:

\begin{enumerate}
\item \textbf{DischargeNet}: This is the Neural Network that is trained to produce a Discharging profile. One DischargeNet model is trained for each battery parameter (DischargNet-V, DischargeNet-T and DischargeNet-S)
\item \textbf{ChargeNet}: This is the Neural Network that is trained to produce a Charging profile. For each battery parameter, one ChargeNet model is trained (ChargNet-V, ChargeNet-T and ChargeNet-S)
\end{enumerate}

\begin{figure*}[htbp!]
\centering
\includegraphics[scale=0.93]{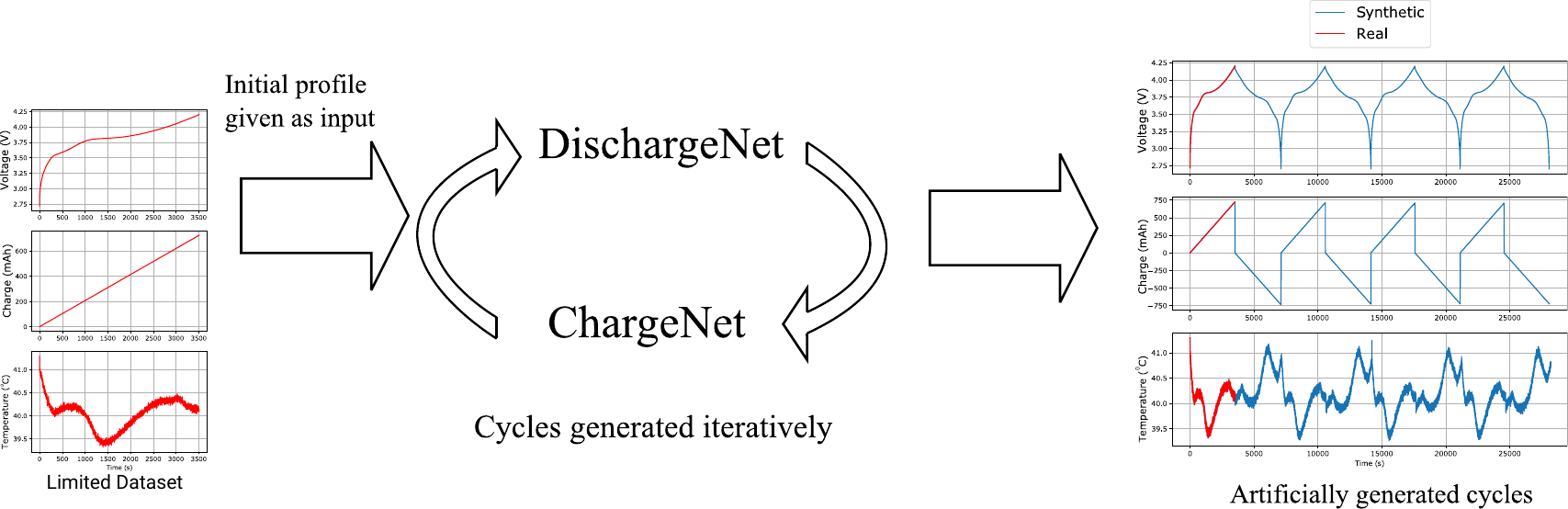}
\caption{Diagram illustrating the methodology followed to produce synthetic cell cycles}
\label{fig:0}
\end{figure*} 

\indent As shown in Fig. \ref{fig:0}, one neural network is trained for the generation of a discharging profile, and one for the generation of the charging profile. The DischargeNet model produces the parameter discharging profile (voltage (V), charge capacity (mAh) or temperature(\degree C)) of a cell. Similarly, a neural network dubbed 'ChargeNet' is trained to produce the charging profile. These models can be used for synthetic data generation, as in the case when only a limited discharging profile data is available, DischargeNet can be used to provide the corresponding discharging profile. In the same way, ChargeNet can be used to produce charging profiles if limited battery charging data is provided to the model.\\

\indent It is possible to use both these models sequentially. This synthetic data generation process can be continued until the accumulated error passes a certain threshold. The net result is to produce a large amount of synthetic data which can be used to develop machine learning -based battery parameter estimation models. This process is elucidated in Fig.~\ref{fig:0}.\\
\indent In this work, we have focused on the Feedforward Neural Network architecture. Although FNNs have been used in the field of Lithium Ion battery parameter estimation before, ours is the first work to propose a concise methodology for sequential synthetic battery cyclic data generation. Previous approaches used a single Neural Network to estimate battery parameters like State-of-Charge and State-of-Health. In this work, we propose a novel methodology of combining two neural networks working together sequentially to supplement existing datasets, with the aim of increasing the amount of data available for training purposes.\\
\indent The architecture of the neural network that is optimal for each of these battery parameters must first be discovered. There are several factors that influence a neural network`s architecture. In the case of the FNN, these are the number of layers and the number of neurons in each layer, which are jointly referred to as the neural network's hyperparameters. This means that there must first be a hyperparameter tuning step before we can proceed to generate artificial data. For each of the battery parameters, we perform this hyperparameter tuning. \\
\indent Our method allows a model that has been trained on data for a specific cell to be used to supplement existing datasets. A model that is trained once for a specific battery chemistry can be used to augment other charging-discharging datasets. For example, if a model can take one cycle and then sequentially generate 50 more cycles artificially, the dataset can essentially be increased 50 times. In this work, we will explore several different FNN architectures for the generation of artificial data on a Kokam Co. Ltd. (SLPB533459H4, 740mAh) cell from the Oxford Battery Degradation Dataset\cite{oxford_dataset}.\\
\indent We first use a limited amount of the full dataset and train several different model architectures for 50 epochs. We do this for the DischargNet models first, and in this work we make the assumption that for a particular battery parameter, the optimal DischargeNet and ChargeNet model architectures are identical. This is a reasonable assumption to make given that both processes are highly similar in terms of internal battery dynamics. 

\section{Dataset}

\begin{figure*}[htbp!]
\centering
\includegraphics[scale=0.43]{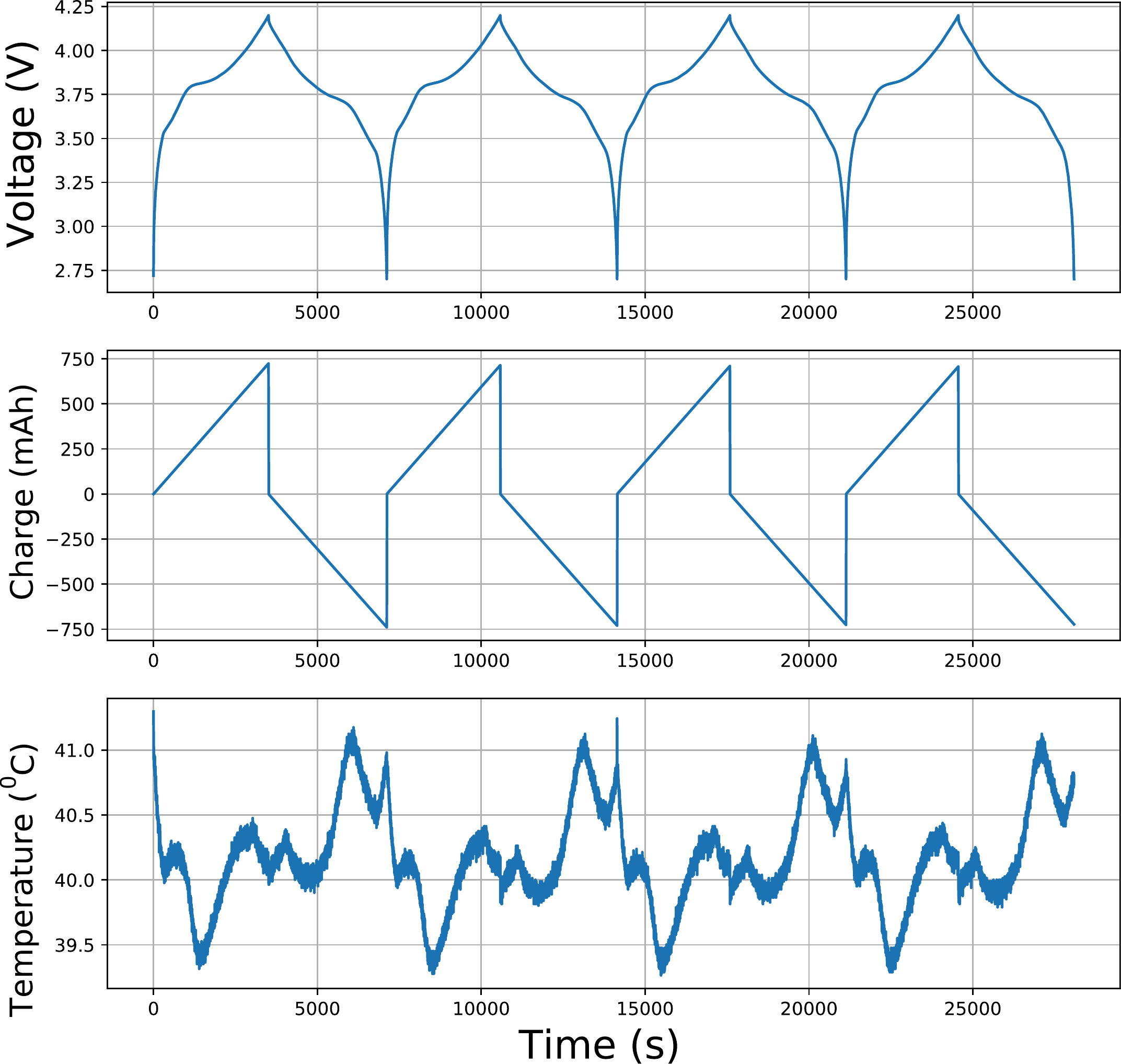}
\caption{Four charge-discharge cycles of Cell 1}
\label{fig:1}
\end{figure*}

\indent The Oxford Battery Degradation dataset \cite{oxford_dataset}$^,$\cite{ChristophBirkl2017} was used in this work. It consists of charging-discharging cycles of eight Kokam Co. Ltd., SLPB533459H4 cells. Of these eight cells, we have used the profiles of the first three cells, henceforth referred to as Cells 1, 2 and 3. The cell specifications are shown in Table 1. The voltage, temperature and charge values of the cells as they were repeatedly charged and discharged were captured using the Bio-Logic MPG-205 battery tester, with a controlled ambient temperature of 40\degree C. There are 78 cycles corresponding to Cell 1, 73 cycles for Cell 2 and 76 cycles for Cell 3. A few of these cycles are shown in Fig. \ref{fig:1}. 

\begin{table*}[t]
\begin{center}
\caption{Cell specifications of Kokam SLPB533459H4 cells\cite{oxford_dataset}}
 \begin{tabular}{c r r }
 \hline
 \textbf{Specification} & \textbf{Value}\\
 \hline
Nominal Capacity & 740 mAh \\
Nominal Voltage & 3.7 V \\
Maximum Voltage & 4.2 V \\
Discharging Cut-off Voltage & 2.7 V\\
Cycle Life & 500 cycles\\
Charging Temp. range & 0\degree C to 40\degree C\\
Discharging Temp. range & -20\degree C to 60\degree C\\
\hline
   \end{tabular}
  \end{center}
\label{table:specs}
\end{table*} 

\indent One of the constraints imposed by the neural network training process in this current research work is that the input and output sequences must have similar dimensions. This means that the time-series sequence given as input must cover the same amount of time as the output time-series. In order to achieve this, a pre-processing step is introduced in order to make sure that the charging and discharging profiles are of equal length. Since the discharging profiles are longer than the charging profiles by roughly 100 time steps, we extend the charge profiles by this amount at the end with the last voltage, charge or temperature value of the series. This ensures that the charging and discharging profiles are of equal length, without introducing unrealistic battery observations into the data. The Center for Advanced Life Cycle Engineering Dataset was also used to test the models for dynamic situations. Two drive cycles (BJDST and US06) sampled at 25\degree C were used from the INR 18650-20R cell dataset\cite{Zheng2016}.

\section{Hyperparameter Tuning}
\subsection{Voltage}

\indent In order to find the best network architecture for this voltage profile modelling task, we must first perform hyperparameter tuning. Initially, different model architectures are trained on only the first two cycles of Cell 1, for 50 epochs, to select the best parameters and loss values, 19553 and  0.00578 were the metrics used in this case, which provide the best accuracy.  
 
\begin{figure*}[htbp!]
\centering
\includegraphics[scale=0.33]{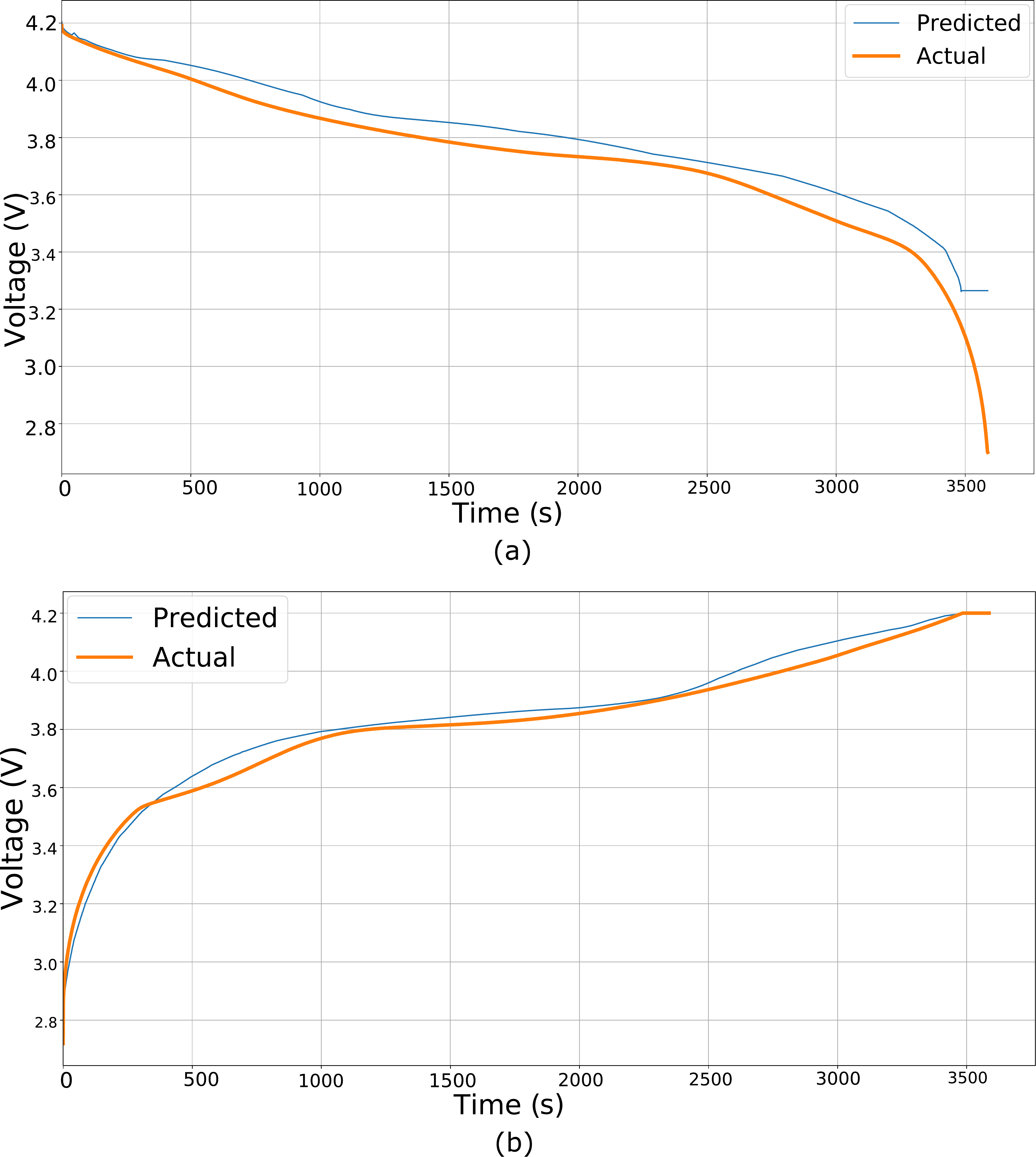}
\caption{DischargeNet output for Cell 3 cycle 1 Voltage profile is shown in (a) ChargeNet output for Cell 3 cycle 1 Voltage profile is shown in (b)}
\label{fig:3_4}
\end{figure*}

The number of parameters help to measure the network complexity. A neural network consists of neurons, each having a specific weight and bias. These weights and biases together constitute the number of trainable parameters of a neural network. Using total number of trainable parameters to gauge network complexity is vital in order to allow researchers to compare the performance of their models with those of other researchers \cite{Vidal2020}. The larger the number of parameters, complexity of the model increases. A deeper network with a large number of layers will have more neurons, and hence more number of parameters.\\
\indent It was observed that, the neural network with 19,553 parameters was the best performing model. This model consisted of 10 layers. We will proceed with training this architecture using the full training dataset for 400 epochs. The training dataset comprises of cycle data of cell 1 and cell 2, a total of 151 charge-discharge cycles. We use cell 3 for testing purposes, which has 76 cycles. For DischargeNet, we train it to produce discharging profiles. All the cycles from Cell 3 are used for testing, and one output is shown in Figs \ref{fig:3_4}. In similar fashion, ChargeNet was trained to predict charging profiles.
%
 
\newpage
\subsection{State-of-Charge}
\begin{figure*}[htbp!]
\centering
\includegraphics[scale=1.1]{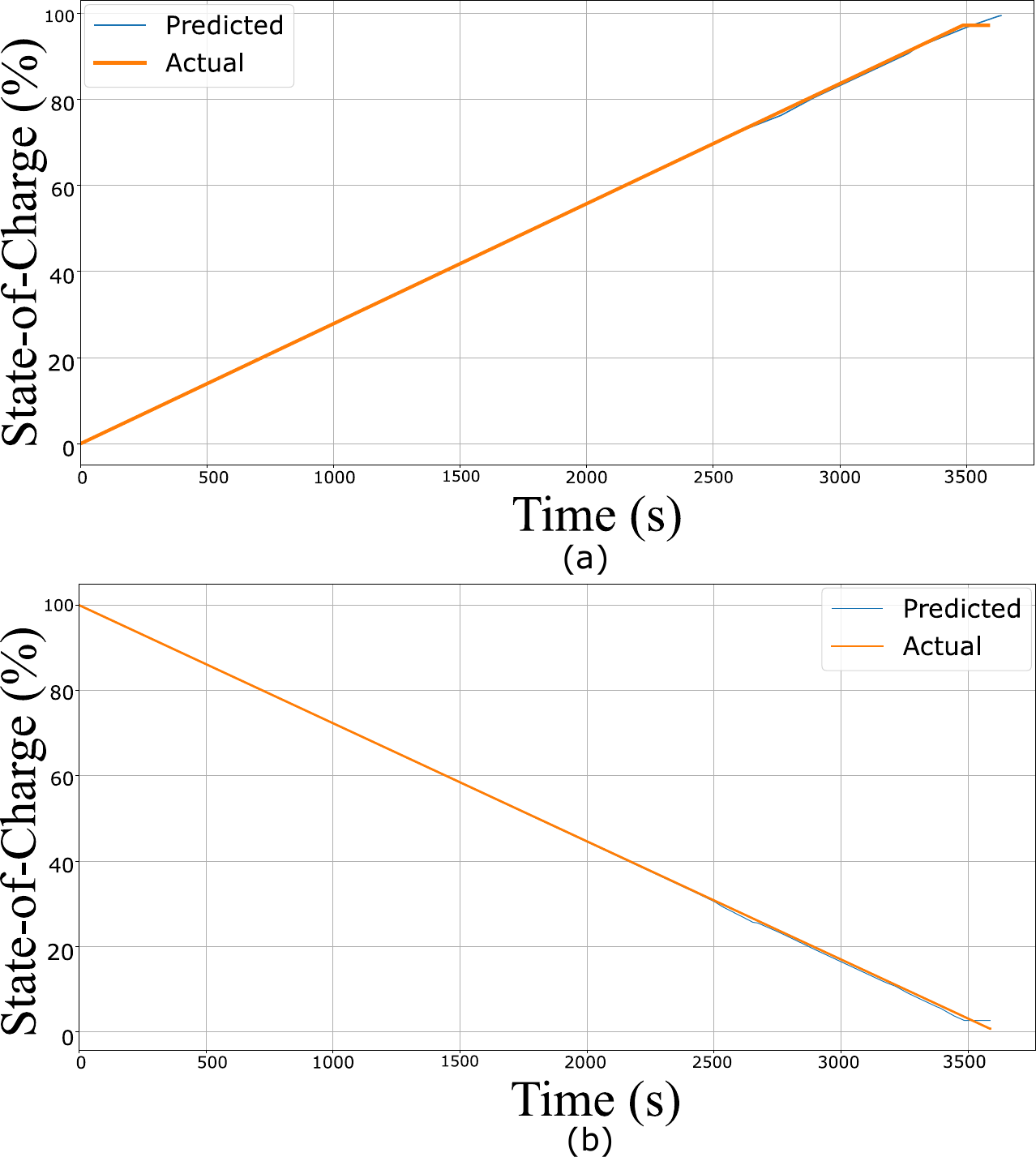}
\caption{ChargeNet output for Cell 3 cycle 1 State-of-Charge profile as shown in (a) and DischargeNet output for Cell 3 cycle 1 State-of-Charge profile as shown in (b)}
\label{fig:5_6}
\end{figure*}

\indent The charge capacity (mAh) values are first converted to State-of Charge values (\%). We repeat the same procedure as that done with the Voltage profile generation for generation of State-of-Charge data. Model with 7,041 parameters, with 6 layers, provided the best performance. The charging and discharging State-of-Charge profiles are shown in Figs. \ref{fig:5_6}. There is marked agreement between the predicted and real datapoints as seen from the plots.


\clearpage

\subsection{Temperature}
\begin{figure*}[b]
\centering
\includegraphics[scale=0.4]{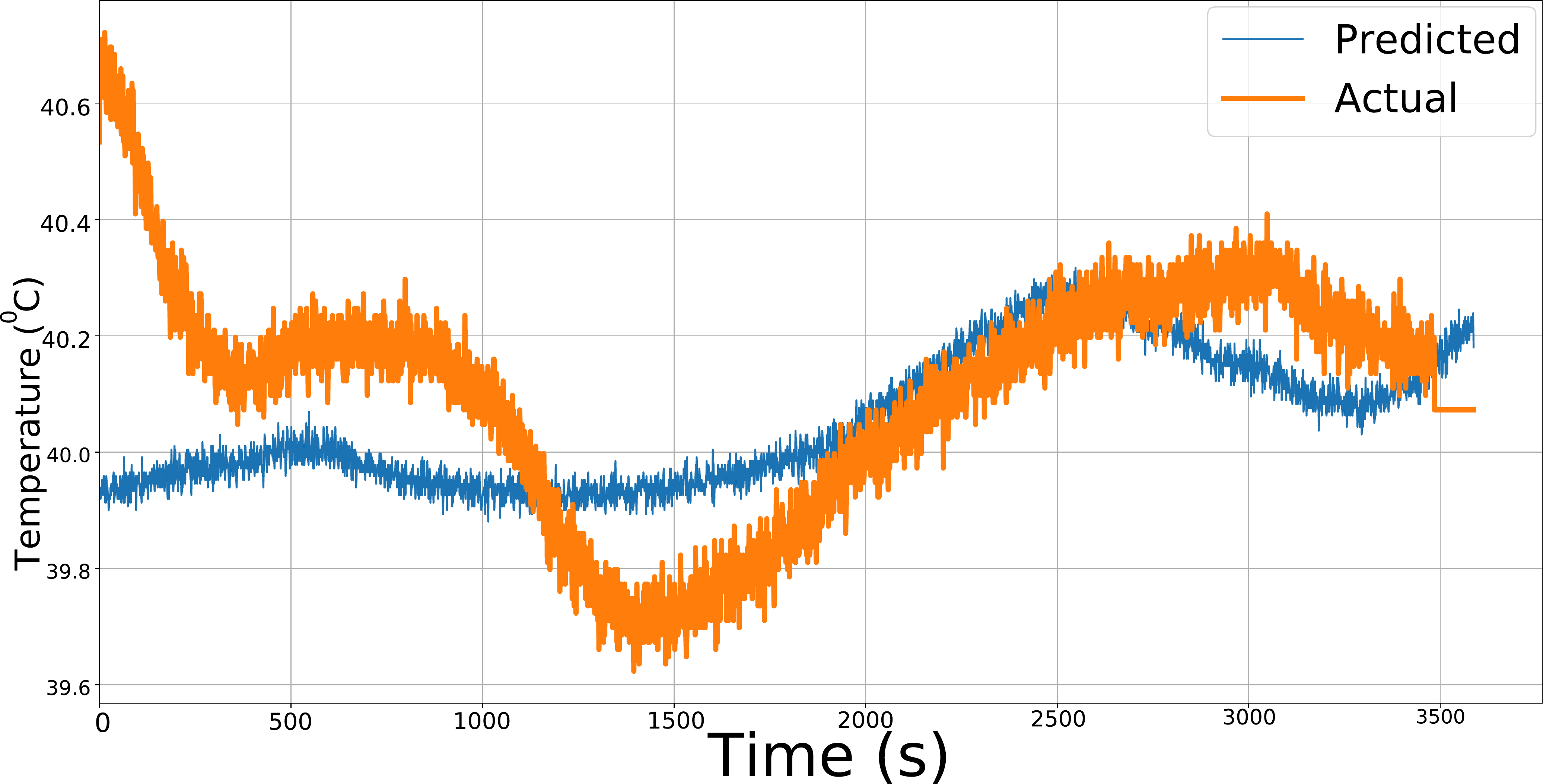}
\caption{ChargeNet output for Cell 3 cycle 1 Temperature profile}
\label{fig:9}
\end{figure*}

\begin{figure*}[htbp!]
\centering
\includegraphics[scale=0.4]{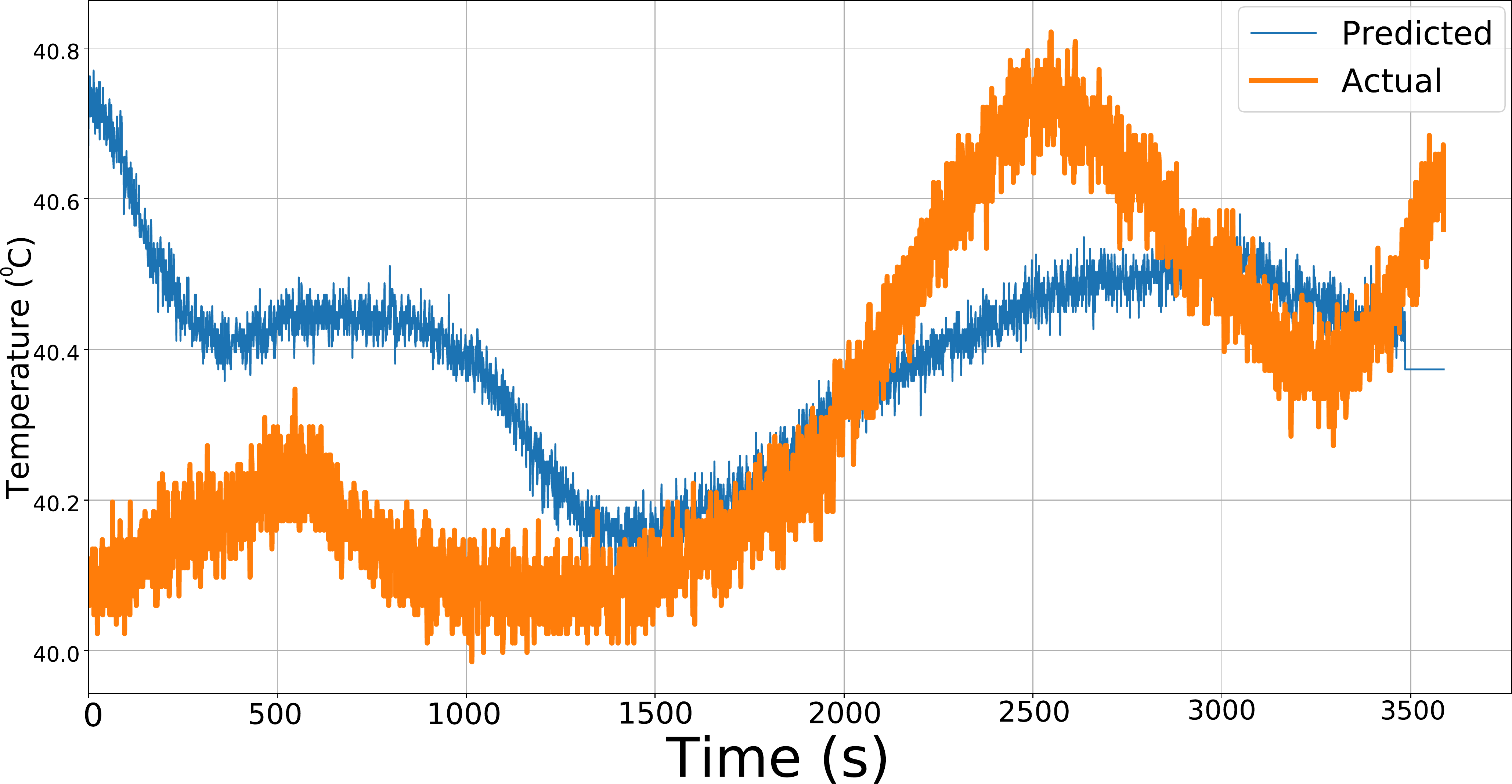}
\caption{DischargeNet output for Cell 3 cycle 1 Temperature profile}
\label{fig:10}
\end{figure*}

We employ the DischargeNet and ChargeNet models to produce synthetic battery temperature data. In this scenario, the model with 7,041 parameters (6 layers), with a final loss value of 0.1247, provided the best performance. The plots are shown for temperature data in Figs. \ref{fig:9} and \ref{fig:10}. The model is able to effectively capture the highly non-linear dynamics of the temperature profile. The plots show the ground-truth values and the values predicted by the model. It must be noted that the variation show a close-up version of the temperature profile, spanning about 0.8\degree C.

\clearpage
\section{Results and Discussion}

\indent We use the Mean Squared Error (MSE), Mean Absolute Error (MAE) and Root Mean Square Error (RMSE) as evaluation metrics for our models. The equations are provided for each of the evaluation metrics. These metrics have been computed for various battery parameters and have been tabulated in Table~\ref{table:5}.
\begin{equation}
    MSE= \frac{1}{n} \Sigma^n _{i=1} (y_i - x_i)^2
\end{equation}

\begin{equation}
    MAE= \frac{1}{n} \Sigma^n _{i=1} |y_i - x_i|
\end{equation}

\begin{equation}
    RMSE= \sqrt{\frac{\Sigma^n _{i=1} (y_i - x_i)^2}{n} }
\end{equation}

where $n$ is the number of data-points, $y_i$ is the predicted value at the $i$-th instance and $x_i$ is the true value at the $i$-th instance.

\begin{table*}[htb!]
\begin{center}
 \begin{tabular}{c r r r r }
 \hline
\textbf{Battery Parameter} & \textbf{Model} & \textbf{MSE} & \textbf{MAE} & \textbf{RMSE} \\ 
\hline
Voltage & ChargeNet-V & 0.0003 & 0.0148 & 0.0177 \\
 
 & DischargeNet-V & 0.0015 & 0.0214 & 0.0385 \\
\hline
State-of-Charge & ChargeNet-S & 0.1542 & 0.1797 & 0.3853 \\
 & DischargeNet-S & 0.1611 & 0.2052 & 0.3953 \\
\hline
Temperature & ChargeNet-T & 0.076 & 0.191 & 0.258 \\
 & DischargeNet-T & 0.070 & 0.208 & 0.259 \\
\hline
   \end{tabular}
  \end{center}
\caption{Table listing the accuracy metrics for Voltage, State-of-Charge and Temperature profile generation}
\label{table:5}
\end{table*}

\indent Table \ref{table:5} shows the average error over the entire 76 cycles of Cell 3 for the voltage profile of the batteries. As can be observed from the results, the average of the error metrics are very low, depicting that the model is able to reconstruct the charging and discharging profiles accurately. The error does not vary significantly over the 76 cycles of Cell 3 for the voltage profile.\\
\indent It must be noted that these small errors showcase a desirable feature of our approach. The errors are small, which illustrates that the model captures the dynamics of the cells, but the small differences between the actual \& predicted values serves to introduce minor variations in the generated data. These variations are vital for any data-augmentation tasks because otherwise there would not be any advantage gained by using a larger dataset which held identical copies of the same battery parameter sequences.\\
\indent We can see in Table \ref{table:5} that the error metrics for the State-of-Charge parameter are also very low. However, there is an upward trend in the error, as the values of MSE, MAE and RMSE increases steadily with the number of cycles. This can be attributed to the capacity fade of the cells. For example, the final charge value of Cell 3 reduces by approximately 25\% from the 1st to the 76th cycle. Note that the predictions show a very good fit between the actual and predicted values. The temperature profile data error averaged over the 76 cycles of Cell 3 are shown in Table \ref{table:5}. The error metrics are within acceptable limits, showcasing the neural networks ability to capture the highly non-linear variation of the temperature profile. It can be seen that a simple coupled neural network approach is good enough to capture all the temporal dynamics of the dataset. The coupled methodology is highly beneficial to produce synthetic charge/discharge data during limited data cases.

\subsection{Dynamic Drive Cycle Profile}

\begin{figure*}[htbp!]
\centering
\includegraphics[scale=0.4]{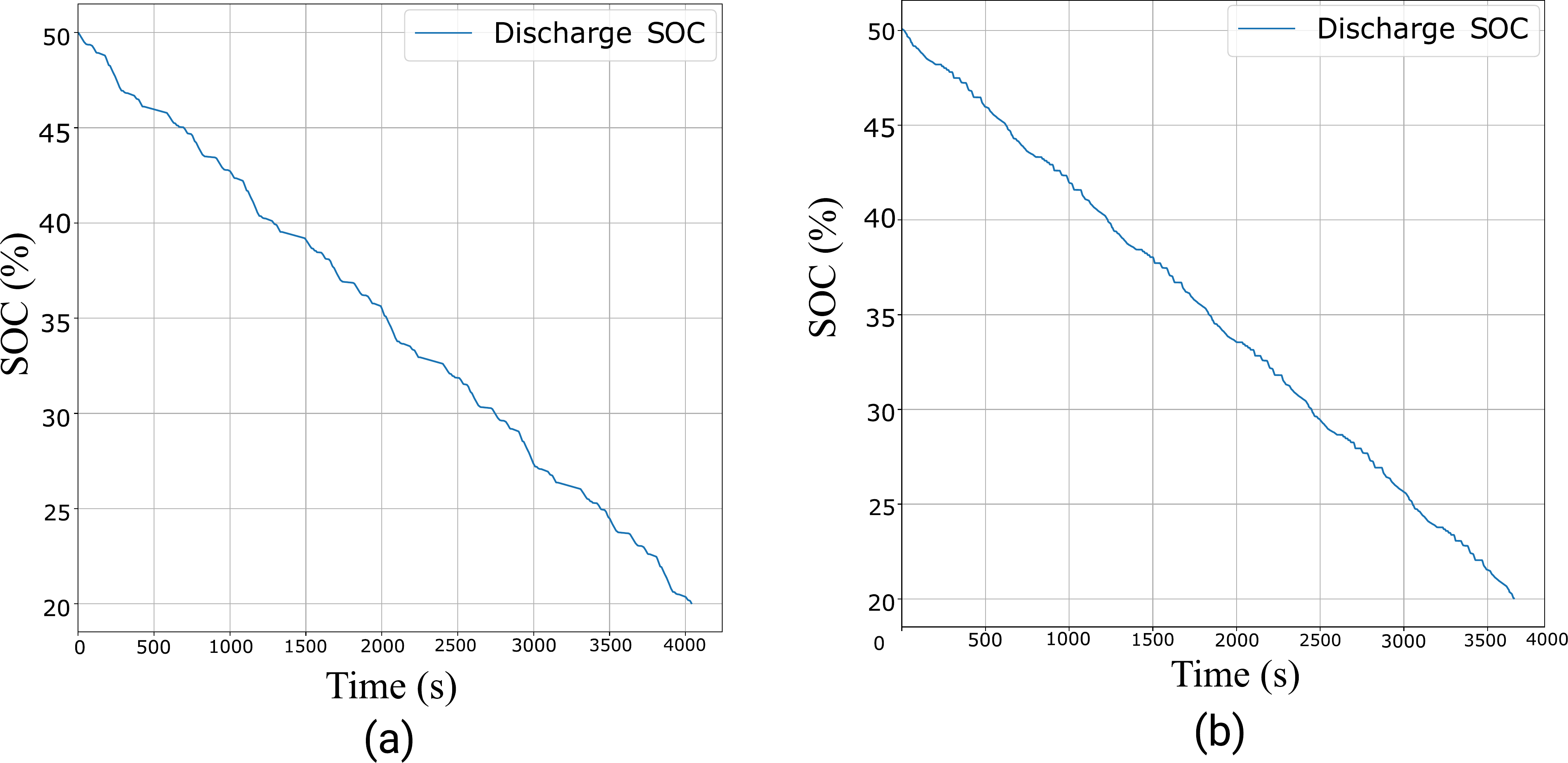}
\caption{Plots describing synthetic dynamic State-of-Charge discharge profiles of BJDST (a) and US06 (b) drive cycle}
\label{fig:17}
\end{figure*}

In order to test our approach on the more challenging scenario of a dynamic drive cycle, the BJDST and US06 drive cycle of an INR 18650-20R cell at 25\degree C is used. These results are shown in Fig. \ref{fig:17}. The State-of-Charge (\%) is synthetically generated using the ChargeNet and DischargeNet model architecture elucidated previously. The model is trained on State-of-Charge data of this cell and the error metrics obtained after testing are shown in Tables \ref{table:8}.
There is an increase in the MAE and RMSE values for these dynamic profiles when compared with simpler charging and discharging battery profiles described in earlier cases. This is due to the transient nature of the battery discharge characteristics which is difficult to model, leading to higher error values. It can be seen that ChargeNet performs better than the DischargeNet model, because charging dynamics are much easier for the Neural Network to capture than the challenging discharging scenario.

\begin{table*}[htbp!]
\begin{center}
 \begin{tabular}{c r r r }
 \hline
\textbf{Drive Cycle} & \textbf{Model} & \textbf{MAE} & \textbf{RMSE} \\ 
\hline
BJDST & ChargeNet & 1.88 & 5.03 \\
 & DischargeNet &  4.81 & 5.91 \\
\hline
US06 & ChargeNet & 2.23 & 2.47 \\
 & DischargeNet &  8.16 & 10.76 \\
\hline 
   \end{tabular}
  \end{center}
\caption{Table depicting accuracy metrics for State-of-Charge battery profile generation for BJDST and US06 drive cycles}
\label{table:8}
\end{table*}

\section{Conclusion}

In this work, we have proposed the use of two coupled neural networks to produce synthetic battery parameter cyclic data to overcome limited dataset challenges. We first perform hyperparameter tuning using limited battery data, and then train the best neural network architecture on the complete dataset. The fully trained networks work together to generate synthetic charge/discharge data for battery parameter values such as Voltage, State-of-Charge and Temperature. It was seen that the Feedforward Neural Network algorithm used in this work was successful in capturing the complete battery dynamics during each cycle, thereby proving the efficacy of this method for data-augmentation purposes. This method was also tested for challenging scenarios of dynamic discharge drive profiles to evaluate the performance of the proposed approach. This is the first work to illustrate the efficacy of simple coupled neural network approach for synthetic data generation which is advantageous for resource constrained devices. Further work can be done to explore the use of Recurrent Neural Networks like Long Short-Term Neural Networks for this task, as these are specifically designed to capture temporal dynamics.


\begin{thebibliography}{50}

\bibitem{Stampatori2020} Stampatori, D., Raimondi, P. P., \& Noussan, M. (2020). Li-ion batteries: A review of a key technology for transport decarbonization. Energies, 13(10). https://doi.org/10.3390/en13102638


\bibitem{Lee2020} Lee, Y., He, L., \& Shin, K. G. (2020). Causes and fixes of unexpected phone shutoffs. MobiSys 2020 - Proceedings of the 18th International Conference on Mobile Systems, Applications, and Services, 206-219. https://doi.org/10.1145/3386901.3389024

\bibitem{Singh2020} Singh, M., Trivedi, J., Maan, P., \& Goyal, J. (2020). Smartphone battery state-of-charge (SoC) estimation and battery lifetime prediction: State-of-art review. Proceedings of the Confluence 2020 - 10th International Conference on Cloud Computing, Data Science and Engineering, 94-101. https://doi.org/10.1109/Confluence47617.2020.9057951


\bibitem{Shen2020}Shen, S., Sadoughi, M., Li, M., Wang, Z., \& Hu, C. (2020). Deep convolutional neural networks with ensemble learning and transfer learning for capacity estimation of lithium-ion batteries. Applied Energy, 260(July 2019), 114296. https://doi.org/10.1016/j.apenergy.2019.114296

\bibitem{Shen2019b} Shen, S., Sadoughi, M., Chen, X., Hong, M., \& Hu, C. (2019). A deep learning method for online capacity estimation of lithium-ion batteries. Journal of Energy Storage, 25(June), 100817. https://doi.org/10.1016/j.est.2019.100817

\bibitem{Venugopal2019} Venugopal, P., \& Vigneswaran, T. (2019). State-of-health estimation of Li-ion batteries in electric vehicle using InDRNN under variable load condition. MDPI Energies, 12(22). https://doi.org/10.3390/en12224338

\bibitem{El-Dalahmeh2020}El-Dalahmeh, M. ;, Al-Greer, M., El-Dalahmeh, M., \& Short, M. (2020). Time-Frequency Image Analysis and Transfer Learning for Capacity Prediction of Lithium-Ion Batteries. MDPI Energies.

\bibitem{Khumprom2019} Khumprom, P., \& Yodo, N. (2019). A data-driven predictive prognostic model for lithium-ion batteries based on a deep learning algorithm. Energies, 12(4). https://doi.org/10.3390/en12040660


\bibitem{Hannan2020} Hannan, M. A., Lipu, M. S. H., Hussain, A., Ker, P. J., Mahlia, T. M. I., Mansor, M., Dong, Z. Y. (2020). Toward Enhanced State of Charge Estimation of Lithium-ion Batteries Using Optimized Machine Learning Techniques. Scientific Reports, 10(1), 1-15. https://doi.org/10.1038/s41598-020-61464-7


\bibitem{Khalid2019} Khalid, A., Sundararajan, A., Acharya, I., \& Sarwat, A. I. (2019). Prediction of Li-Ion Battery State of Charge Using Multilayer Perceptron and Long Short-Term Memory Models. ITEC 2019 - 2019 IEEE Transportation Electrification Conference and Expo. https://doi.org/10.1109/ITEC.2019.8790533

\bibitem{Wassiliadis2019} Wassiliadis, N., Herrmann, T., Wildfeuer, L., Reiter, C., \& Lienkamp, M. (2019). Comparative Study of State-Of-Charge Estimation with Recurrent Neural Networks. ITEC 2019 - 2019 IEEE Transportation Electrification Conference and Expo. https://doi.org/10.1109/ITEC.2019.8790597

\bibitem{Anton2013} Alvarez Anton, J. C., Garcia Nieto, P. J., Blanco Viejo, C., \& Vilan Vilan, J. A. (2013). Support vector machines used to estimate the battery state of charge. IEEE Transactions on Power Electronics, 28(12), 5919-5926. https://doi.org/10.1109/TPEL.2013.2243918

\bibitem{Vidal2020} Vidal, C., Malysz, P., Kollmeyer, P., \& Emadi, A. (2020). Machine Learning Applied to Electrified Vehicle Battery State of Charge and State of Health Estimation: State-of-the-Art. IEEE Access, 8, 52796-52814. https://doi.org/10.1109/ACCESS.2020.2980961

\bibitem{Chemali2018}Chemali, E., Kollmeyer, P. J., Preindl, M., Ahmed, R., \& Emadi, A. (2018). Long Short-Term Memory Networks for Accurate State-of-Charge Estimation of Li-ion Batteries. IEEE Transactions on Industrial Electronics, 65(8), 6730-6739. https://doi.org/10.1109/TIE.2017.

\bibitem{How2019a}How, D. N. T., Hannan, M. A., Lipu, M. S. H., Sahari, K. S. M., Ker, P. J., \& Muttaqi, K. M. (2019). State-of-Charge Estimation of Li-ion Battery in Electric Vehicles: A Deep Neural Network Approach. 2019 IEEE Industry Applications Society Annual Meeting, IAS 2019, 1-8. https://doi.org/10.1109/IAS.2019.8912003


\bibitem{Tran2020a} Tran, M. K., Mevawala, A., Panchal, S., Raahemifar, K., Fowler, M., \& Fraser, R. (2020). Effect of integrating the hysteresis component to the equivalent circuit model of Lithium-ion battery for dynamic and non-dynamic applications. Journal of Energy Storage, 32(August), 101785. https://doi.org/10.1016/j.est.2020.101785

\bibitem{Lai2020a} Lai, X., Wang, S., Ma, S., Xie, J., \& Zheng, Y. (2020). Parameter sensitivity analysis and simplification of equivalent circuit model for the state of charge of lithium-ion batteries. Electrochimica Acta, 330. https://doi.org/10.1016/j.electacta.2019.135239

\bibitem{Marquis2019} Marquis, S. G., Sulzer, V., Timms, R., Please, C. P., \& Chapman, S. J. (2019). An asymptotic derivation of a single particle model with electrolyte. Journal of The Electrochemical Society, 1-43. https://doi.org/10.1149/2.0341915jes


\bibitem{Dubarry2020}Dubarry, M., \& Beck, D. (2020). Benchmark Synthetic Training Data for Artificial 2 Intelligence-based Li-ion Diagnosis and Prognosis. Journal of Power Sources, 479(May), 1-20. https://doi.org/10.1016/j.jpowsour.2020.228806

\bibitem{Pyne2019} Pyne, M., Yurkovich, B. J., \& Yurkovich, S. (2019). Generation of Synthetic Battery Data with Capacity Variation. CCTA 2019 - 3rd IEEE Conference on Control Technology and Applications, 476-480. https://doi.org/10.1109/CCTA.2019.8920488

\bibitem{Kamthe2021}Kamthe, S., Assefa, S., \& Deisenroth, M. (2021). Copula Flows for Synthetic Data Generation. Retrieved from http://arxiv.org/abs/2101.00598



\bibitem{Goodfellow2014} Goodfellow, I. J., Pouget-Abadie, J., Mirza, M., Xu, B., Warde-Farley, D., Ozair, S., Bengio, Y. (2014). Generative adversarial nets. Advances in Neural Information Processing Systems, 3(January), 2672-2680.

\bibitem{DosSantosTanaka2019} dos Santos Tanaka, F. H. K., \& Aranha, C. (2019). Data Augmentation Using GANs. Proceedings of Machine Learning Research 2019, 2019, 1-16.

\bibitem{Chatziagapi2019} Chatziagapi, A., Paraskevopoulos, G., Sgouropoulos, D., Pantazopoulos, G., Nikandrou, M., Giannakopoulos, T., Narayanan, S. (2019). Data augmentation using GANs for speech emotion recognition. Proceedings of the Annual Conference of the International Speech Communication Association, INTERSPEECH, 2019-September, 171-175. https://doi.org/10.21437/Interspeech.2019-2561

\bibitem{Haradal2018} Haradal, S., Hayashi, H., \& Uchida, S. (2018). Biosignal Data Augmentation Based on Generative Adversarial Networks. Proceedings of the Annual International Conference of the IEEE Engineering in Medicine and Biology Society, EMBS, 2018-July, 368-371. https://doi.org/10.1109/EMBC.2018.8512396

\bibitem{Yoon2019} Yoon, J., Jarrett, D., \& van der Schaar, M. (2019). Time-series generative adversarial networks. Advances in Neural Information Processing Systems, 32(NeurIPS), 1-11.

\bibitem{Shorten2019} dos Santos Tanaka, F. H. K., \& Aranha, C. (2019). Data Augmentation Using GANs. Proceedings of Machine Learning Research 2019, 2019, 1-16.

\bibitem{Wu2018} Wu, E., Wu, K., Cox, D., \& Lotter, W. (2018). Conditional infilling GANs for data augmentation in mammogram classification. Lecture Notes in Computer Science (Including Subseries Lecture Notes in Artificial Intelligence and Lecture Notes in Bioinformatics), 11040 LNCS, 98-106. https://doi.org/10.1007/978-3-030-00946-5-11
\bibitem{Kingma2015} Kingma, D. P., \& Ba, J. L. (2015). Adam: A method for stochastic optimization. 3rd International Conference on Learning Representations, ICLR 2015 - Conference Track Proceedings, 1-15.


\bibitem{oxford_dataset} Birkl, C. (2017). Oxford Battery Degradation Dataset 1. University of Oxford.

\bibitem{ChristophBirkl2017} Christoph Birkl. (2017). Diagnosis and prognosis of degradation in lithium-ion batteries. PhD Thesis, University of Oxford. Retrieved from https://ora.ox.ac.uk/objects/uuid:7d8ccb9c-1469-4209-9995-5871fc908b54

\bibitem{Zheng2016} Zheng, F., Xing, Y., Jiang, J., Sun, B., Kim, J., \& Pecht, M. (2016). Influence of different open circuit voltage tests on state of charge online estimation for lithium-ion batteries. Applied Energy, 183, 513-525. https://doi.org/10.1016/j.apenergy.2016.09.010

\end{thebibliography}
\end{document}